\documentclass[acmlarge]{acmart}

\renewcommand\footnotetextcopyrightpermission[1]{} 
\AtBeginDocument{%
  \providecommand\BibTeX{{%
    \normalfont B\kern-0.5em{\scshape i\kern-0.25em b}\kern-0.8em\TeX}}}

\acmJournal{TKDD}

\usepackage{multirow}
\usepackage{graphicx}
\usepackage{subfigure}
\usepackage{threeparttable}
\usepackage{algorithm}
\usepackage{algpseudocode}
\usepackage{bm}
\usepackage{caption}
\usepackage{makecell}
\usepackage{algorithmicx}
\usepackage{url}

\usepackage{color}
\usepackage{enumitem}
\usepackage[skip=5pt]{caption}
\usepackage{booktabs}
\newcommand{\secref}[1]{Section \ref{#1}}
\newcommand{\figref}[1]{Figure \ref{#1}}
\newcommand{\eqnref}[1]{Eq. (\ref{#1})}
\newcommand{\tabref}[1]{Table \ref{#1}}

\begin{document}

\title{FMGNN: Fused Manifold Graph Neural Network}


\author{Cheng Deng}
\author{Fan Xu}
\author{Jiaxing Ding}
\author{Luoyi Fu}
\author{Weinan Zhang}
\author{Xinbing \\Wang}
\email{davndw@sjtu.edu.cn}
\affiliation{%
  \institution{Shanghai Jiao Tong University}
  \city{Shanghai}
  \country{China}}

\author{Chenghu Zhou}
\affiliation{%
  \institution{Institute of Geographical Science and Natural Resources Research, Chinese Academy of Sciences}
  \city{Beijing}
  \country{China}}
\email{zhouch@lreis.ac.cn}

\renewcommand{\shortauthors}{Cheng Deng, et al.}

\begin{abstract}
    Graph representation learning has been widely studied and demonstrated effectiveness in various graph tasks. Most existing works embed graph data in Euclidean space, while recent works extend the embedding models to hyperbolic or spherical spaces to achieve better performance on graphs with complex structures, such as hierarchical or ring structures. Fusing the embedding from different manifolds can take advantage of the embedding capabilities over different graph structures. However, existing embedding fusion methods mainly focus on concatenating or summing up the output embeddings without considering interacting and aligning the embeddings of the same vertices on different manifolds, which can lead to distortion and imprecision in the final fusion results. Besides, it is also challenging to fuse the embeddings of the same vertices from different coordinate systems. In the face of these challenges,  we propose the \textbf{F}used \textbf{M}anifold \textbf{G}raph \textbf{N}eural \textbf{N}etwork (\textbf{FMGNN}). This novel GNN architecture embeds graphs into different Riemannian manifolds with interaction and alignment among these manifolds during training and fuses the vertex embeddings through the distances on different manifolds between vertices and selected landmarks, geometric coresets. Our experiments demonstrate that FMGNN yields superior performance over strong baselines on the benchmarks of node classification and link prediction tasks.
\end{abstract}





\maketitle

\section{Introduction}
The real world graphs always exist with inherent complex structures and graph representation learning models, such as Graph Neural Network (GNN)~\cite{scarselli2008graph}, and Graph Convolutional Network (GCN)~\cite{kipf2016semi}, have demonstrated effectiveness in embedding graphs into vector spaces
to capture complex relations between vertices. 

However, some peculiar but widespread structures, such as hierarchical or ring, that can not be well preserved by Euclidean space~\cite{sarkar2011low,Zhu2020GraphGI},
and this leads to researchers' studies on the non-Euclidean space. 
Hyperbolic space has been introduced~\cite{chami2019hyperbolic,liu2019hyperbolic} to embed graphs  with scale-free or hierarchical structures,  achieving better performance than that on the Euclidean spaces~\cite{nickel2017poincare}.  
The hyperbolic space can manage to preserve the distance relations between vertices on hierarchical graphs, whose neighborhood sizes grow exponentially with the depth of the hierarchy~\cite{Neighborhood}, where the Euclidean space fails. 
In addition, spherical space is studied~\cite{xiao2015one,Bachmann2020ConstantCG} to improve
embedding for graphs with ring structures. 
Different spaces can be adopted to improve the embedding performance of graphs with particular structures. 
However, these works assume that all the vertices are embedded on the space with the same curvature, which is not general enough to fit real world graphs with various complex structures. 

In order to suit the graphs with a variety of topology, it is intuitive to combine the embedding from different curvatures and spaces into a unified embedding paradigm.~\cite{gu2018learning,Zhu2020GraphGI} 
Nevertheless, these works either concatenate or sum up the final obtained embeddings from different spaces, without take the interaction and alignment of vertex embeddings of different layers into consideration during training process. Consequently, once the embeddings in different spaces change in completely different directions during the training process, the final obtained embedding may be distorted and variant.
As different vertices can have different best-fit spaces to preserve the relations on the graph~\cite{Zhu2020GraphGI}, a vertex's embedding can benefit from the aggregated information from its neighbors in other spaces during message passing, if the neighbors can be best embedded in the other spaces. This kind of interaction and aggregation can be operated by the geometry interaction over manifolds. Nevertheless, without proper interaction and alignment, the embeddings of same vertices obtained from different spaces can be of random shift and distortion due to randomness during model training, even though the mutual relative relations between vertices in each respective space are preserved at best. 
Thus, concatenating or mixing such embeddings results in imprecision and fluctuation, which we will demonstrate in this work.  
Moreover, it would be difficult to reasonably fuse the embeddings and design further operations, such as regression and softmax~\cite{liu2019hyperbolic}, since such operations on the coordinates of different spaces are different. 

\begin{figure}[ht]
  \centering
  \includegraphics[width=\linewidth]{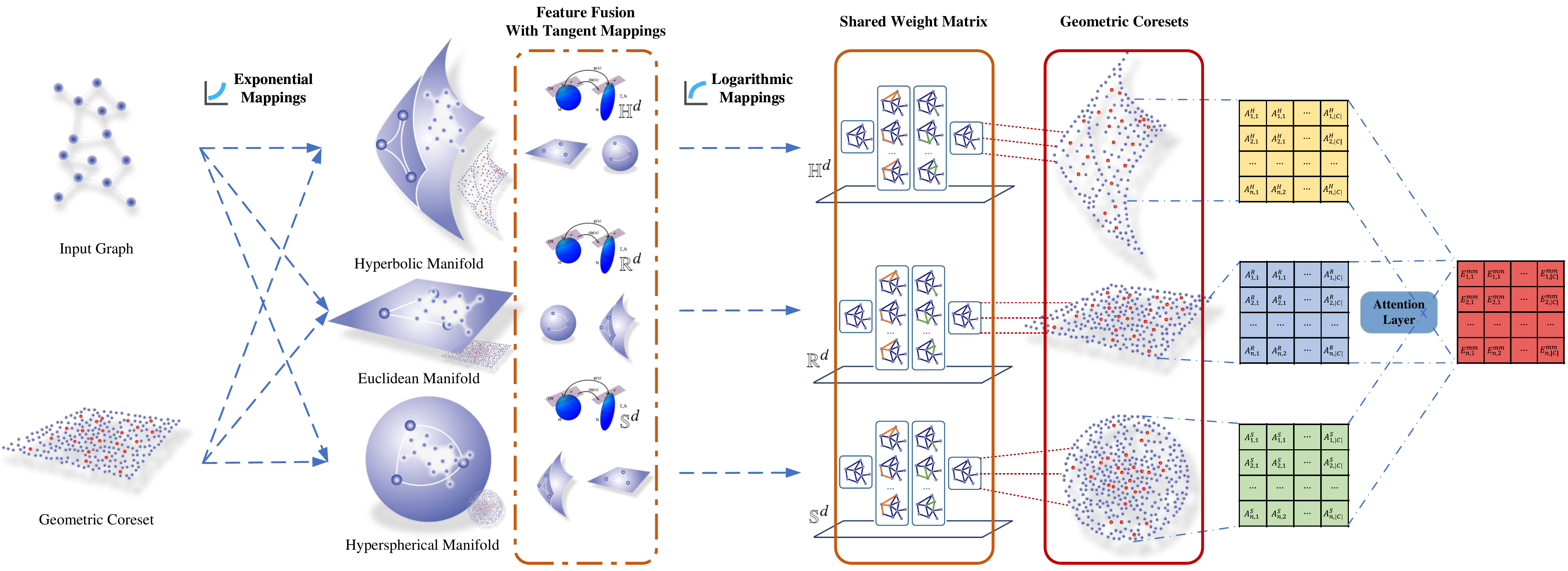}
  \caption{Overview of the Fused Manifold Graph Neural Network (FMGNN). In FMGNN, a graph will be embedded via GCNs on three Riemannian manifolds, manifold fusion via tangent mapping before neighborhood aggregation, information aggregation on each manifold by distances to landmark geometry coresets, and adjust to the final representation with a self-attention mechanism.}
  \label{fig:1}
  \vspace{-1em}
\end{figure}

To this end, in this work, we propose Fused Manifold Graph Neural Network (\textbf{FMGNN}), which properly aligns embeddings on different manifolds with interations via tangent mapping before neighborhood aggregation and fuses information on each manifold by distances to landmark geometry coresets.
With \textbf{FMGNN}, embeddings from different manifolds are appropriately aligned during training and efficiently fused in the final results.
The overview flowwork of FMGNN is illustrated in \autoref{fig:1}. 
While elaborating the details in~\secref{sec:fmgnn}, for ease of understanding, here we would like to briefly unfold the cores components of FMGNN. We first utilize the exponential and logarithmic mappings to bridge the spherical, Euclidean, hyperbolic manifold embeddings and propose the tangent mapping for the interaction among the vertex embeddings on different manifolds. In this way, the embedding of the same vertices are properly aligned and the information of the neighborhood on different manifolds are well communicated. 
When it comes to fuse the final embedding obtained from different manifolds, we turn vertex embedding into the features measured by the distance to the centroid coreset. Hence, FMGNN still guarantees near vertices can have similar distances, a figuration of coordinates. 


Our contributions are summarized as follows:

\begin{itemize}[leftmargin=0.8em]
    \item We propose a novel graph representation learning framework, fused manifold graph neural network (FMGNN), which, to the best of our knowledge, first studies graph embedding on all the 
    Euclidean, hyperbolic, and spherical manifolds, with novel fusion method based on tangent mapping to align embeddings from different spaces and integrate neighborhood information throughout each embedding layer during training. 
    \item We put forward a novel geometric coreset-based centroid mapping method, which can find the coreset of nodes from randomly sampled nodes in manifolds, and unify the graph embedding of different manifolds into pairwise distance between the vertices and the geometric coresets on the corresponding manifolds. 
    \item Extensive experiment results show that FMGNN has achieved SOTA performance on several benchmarks, 
    especially including graphs with sparsity and high hyperbolicity. Our model absorbs the advantages of hyperbolic models on tree structure and maintains the strength of Euclidean space in some free scale topology.
\end{itemize}

The rest of the paper is organized as follows: In Section 2, we list the related works of FMGNN. In Section 3, the background, metrics, operational rules of Riemannian manifolds are introduced. Particularly, we present the observation on GNN training over different manifolds in Section 4. Then, we share the methodology of FMGNN in Section 5. Finally, the details of experiments on FMGNN will be discussed in Section 6.


\section{Related Work}

In this section, we briefly review the related works about GNNs on both Euclidean and non-Euclidean manifolds, along with the geometric coreset. 

\textbf{GNN on the Euclidean manifold.} GNNs has received rising attention in the field of machine learning since it was first put forward~\cite{gori2005new,scarselli2008graph}, especially after the  GCN~\cite{kipf2016semi} was proposed, which combines the information of neighbor nodes with the message passing mechanism. Further, the GAT~\cite{velivckovic2017graph} introduces the attention mechanism into GCN and the SGC~\cite{wu2019simplifying} convert the nonlinear GCN into a simple linear model. For large-scale graphs, with the high computation cost of unified message passing, GraphSAGE~\cite{hamilton2017inductive} is proposed on node sampling and hierarchical sampling. 
In complex network analysis, topology on edges have different effects of information dissemination, CurvGN~\cite{ye2019curvature} incorporates Ricci curvature to measure the topology difference into the message passing as an important parameter in the attention mechanism of GCN, while M$_{2}$GNN~\cite{Wang2021MixedCurvatureMG} set curvature as a learnable parameter to train knowledge graph embedding.

\textbf{GNN on the non-Euclidean manifold.} GNN on the non-Euclidean manifold has attracted more and more attention recently. 
Texts are embedded into the hyperbolic or spherical manifold for better performance with word2vec~\cite{xiao2015one,meng2019spherical,dhingra2018embedding}, and the multiple relations in the knowledge graph are also successfully embedded in the hyperbolic space~\cite{chami2020low}. 
The  HNN~\cite{ganea2018hyperbolic} trains neural network in the hyperbolic space, in the tasks of the parametric classification and regression methods. 
HGCN~\cite{chami2019hyperbolic} and HGNN~\cite{liu2019hyperbolic} embed graphs into hyperbolic space and apply convolution computation on hyperbolic vector mapping and transition. Recently, Geometric Interaction Learning (GIL)~\cite{Zhu2020GraphGI} put forward a schema to fuse features at the last layer of the GCN on different manifolds, while $\kappa$-GCN~\cite{Bachmann2020ConstantCG} investigate the constant curvature GNN finding that different datasets suit different manifolds. In addition, H2H-GCN~\cite{Dai_2021_CVPR} proposes the direct convolution mechanism without the transformation via the tangent space in a hyperbolic manifold. 
What's more, a theoretical analysis on the suitable curvature space for a real world graph is conducted~\cite{Neighborhood}. And GIE~\cite{cao2022geometry} deploys a similar mechanism as GIL~\cite{Zhu2020GraphGI} on knowledge graph embedding scenario.

\textbf{Geometric Coreset.} In our work, we make use of the geometric coreset~\cite{bachem2017scalable,sener2017active}, which is widely used to approximate the geometry properties with a constant size of points, in the field of computational geometry, statistics and computer vision. 
In geometry deep learning, geometric coreset is used to draw the outline of the geometry elements efficiently with the constant number of points~\cite{sinha2020small} and analyzing the coresets or sub-level sets of the distance to a compact manifold is a common method in topology data analysis to understand its topology~\cite{Brcheteau2018TheK}. Moreover, the paradigm of coresets has emerged as an appropriate means for efficiently approximating various extent measures of a point sets~\cite{Agarwal2007GeometricAV}. This inspires us to adopt coreset-related means to measure the features of the nodes on graph 
so that the nodes can be represented with a relatively low-dimension embedding.
As mentioned and discussed in HGNN~\cite{liu2019hyperbolic}, regression and classification methods can not suit well on non-Euclidean manifolds. HGNN uses a set of coordinate-trainable nodes as reference frame to transform the nodes coordinates into nodes features and these features represents the node-wise distance with the reference nodes.

FMGNN, proposed in this paper, tries to avoid downgrading and unstable performances by deploy tangent mapping before neighborhood aggregation to align the embedding (coordinates) from Euclidean and non-Euclidean manifolds, and take coreset as a heuristic ``landmark'' to extract and fuse information on each manifold by distances.

\section{Preliminaries}
In this section, we will recall the metrics and operational rules on Riemannian manifolds~\cite{willmore2013introduction}. 

\begin{table}[h]
  \centering
  \begin{tabular}{lccc}
  \hline
   & Euclidean $\mathbb{R}^{d}$& Hyperbolic $\mathbb{H}^{d}$ & Spherical $\mathbb{S}^{d}$\\ \hline
  Curvature &     $\kappa = 0$     &   $\kappa < 0$   &  $\kappa > 0$   \\
  Exponential Mapping & $\mathbf{x}+\mathbf{v}$ & $\cosh \left(\|\mathbf{v}\|_{\mathbb{H}}\right) \mathbf{x}+\sinh \left(\|\mathbf{v}\|_{\mathbb{H}}\right) \frac{\mathbf{v}}{\|\mathbf{v}\|_{\mathbb{H}}}$ & $\cos \left(\|\mathbf{v}\|_{\mathbb{S}}\right) \mathbf{x}+\sin \left(\|\mathbf{v}\|_{\mathbb{S}}\right) \frac{\mathbf{v}}{\|\mathbf{v}\|_{\mathbb{S}}}$\\ 
  Logarithmic Mapping & $\mathbf{y}-\mathbf{x}$ & $d_{\mathbb{H}}(\mathbf{x}, \mathbf{y}) \frac{\mathbf{y}+\langle\mathbf{x}, \mathbf{y}\rangle_{\mathbb{H}} \mathbf{x}}{\left\|\mathbf{y}+\langle\mathbf{x}, \mathbf{y}\rangle_{\mathbb{H}} \mathbf{x}\right\|_{\mathbb{H}}}$ & $d_{\mathbb{S}}(\mathbf{x}, \mathbf{y}) \frac{\mathbf{y}-\langle\mathbf{x}, \mathbf{y}\rangle_{\mathbb{S}} \mathbf{x}}{\left\|\mathbf{y}-\langle\mathbf{x}, \mathbf{y}\rangle_{\mathbb{S}} \mathbf{x}\right\|_{\mathbb{S}}}$ \\ \hline
  \end{tabular}
  \caption{Exponential and logarithmic Mappings of the Riemannian manifolds. ($\mathbb{R},\mathbb{H},\mathbb{S}$).}
  \label{tab:manifold}
  \vspace{-1em}
\end{table}

\subsection{Riemannian Manifolds}
A $d$-dimension Riemannian manifold $\mathcal{M}^{d}$ is a differential manifold equipped with an inner product on tangent space $g_{\mathbf{x}}: \mathcal{T}_{\mathbf{x}} \mathcal{M}^{d} \times \mathcal{T}_{\mathbf{x}} \mathcal{M}^{d} \rightarrow \mathbb{R}$ at each point $\mathbf{x} \in \mathcal{M}^{d}$ where the tangent space $\mathcal{T}_{\mathbf{x}} \mathcal{M}^{d}$ is a $d$-dimensional vector space, which can be regarded as a flat Euclidean-like space, an approximation of neighborhood of x on the manifold, and its metric changes smoothly with $\mathbf{x}$~\cite{robles2007riemannian}
Specifically, there are three classic Riemannian manifolds: \textbf{Euclidean Manifold $\mathbb{R}^{d}$} is a manifold with zero curvature. 
\textbf{Hyperbolic Manifold $\mathbb{H}^{d}$} is a manifold with the constant negative curvature. There are several canonical models for hyperbolic manifolds, such as the Poincar{\'e} Ball model \cite{nickel2017poincare} and the Lorentz model \cite{nickel2018learning}. In this paper, we adopt Poincar{\'e} ball model to FMGNN since the dimension of Poincar{\'e} ball model is equal to the Euclidean and the spherical manifolds. Hyperbolic manifold shaped by Poincar{\'e} ball model, denoted as $\mathbb{H}^{d}=\left\{\mathbf{x} \in \mathbb{R}^{d+1}:\langle\mathbf{x}, \mathbf{x}\rangle_{\mathbb{H}}=-1, x_{0}>0\right\}$ with the Riemannian metric $g^{\mathbb{H}}=Diag([-1,1, \ldots, 1])$. 
\textbf{Spherical Manifold $\mathbb{S}^{d}$} is a manifold with constant positive curvature. In general, spherical geometry provides benefits for modeling spherical or cyclical data. We adopt the spherical manifold to graph embedding and fuse it with the Euclidean and the hyperbolic manifold, referring to several related works \cite{shin2020robust,yang1957maps}. 
The spherical manifold in 2D is isomorphic to the 2D surface of a spherical embedding in 3D Euclidean space which can be generalized to an n-dimensional spherical embedding in an n-dimensional Euclidean space. 

\subsubsection{Exponential and logarithmic mappings}
Exponential and logarithmic mappings transform the tangent space and manifolds. 
The exponential mapping, $\exp _{\mathbf{x}}: \mathcal{T}_{\mathbf{x}} \mathcal{M}^{d} \rightarrow \mathcal{M}^{d}$, is the map which connects the Lie algebra on tangent space $\mathcal{T}_{\mathbf{x}} \mathcal{M}^{d}$ to the Lie group which defines the manifolds $\mathcal{M}^{d}$. It is defined on $\mathcal{M}^{d}$ and is a far-reaching generalization of the ordinary exponential function. The logarithmic mapping, $\log _{\mathbf{x}}: \mathcal{M}^{d} \rightarrow \mathcal{T}_{\mathbf{x}} \mathcal{M}^{d}$, is the inverse of an exponential mapping, transferring points of a manifold $\mathcal{M}^{d}$ back to the tangent space, such that $\log _{\mathbf{x}}\left(\exp _{\mathbf{x}}(\mathbf{v})\right)=\mathbf{v}, v \in \mathcal{M}^{d}$. A brief illustration is shown in \figref{fig:2}. We directly show the equations for the two mappings in the three spaces in \tabref{tab:manifold}. 

Meanwhile, we remind the plain-vanilla $M\ddot obius$ addition \eqnref{m_add}
~\cite{Bachmann2020ConstantCG} related to a manifold of constant sectional curvature $\kappa$ for further feature fusion:
{\small
\begin{equation}
  \label{m_add}
  \mathbf{x} \oplus_{\kappa} \mathbf{y}=\frac{\left(1-2 \kappa\langle\mathbf{x}, \mathbf{y}\rangle-\kappa\|\mathbf{y}\|_{2}^{2}\right) \mathbf{x}+\left(1+\kappa\|\mathbf{x}\|_{2}^{2}\right) \mathbf{y}}{1-2 \kappa\langle\mathbf{x}, \mathbf{y}\rangle+\kappa^{2}\|\mathbf{x}\|_{2}^{2}\|\mathbf{y}\|_{2}^{2}}.
\end{equation}
}


\subsubsection{Tangent mapping} 
\begin{figure}[h]
  \centering
  \includegraphics[width=0.8\linewidth]{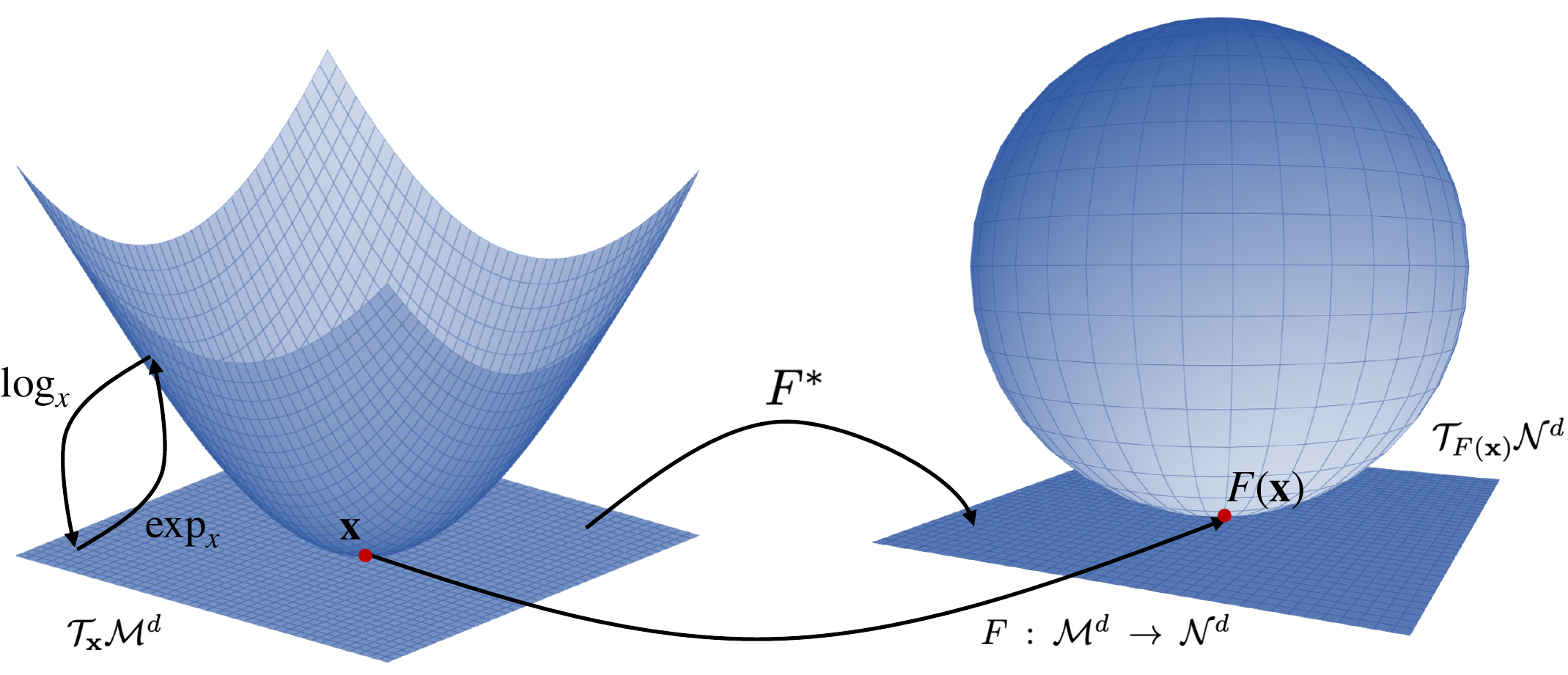}
  \caption{The illustration of exponential (Mapping from tangent spaces to manifolds), logarithmic  (Mapping from manifolds to tangent spaces), and tangent mappings ($F^{*}$) on non-Euclidean manifolds (e.g. Mapping from hyperbolic space to spherical space, $F: \mathcal{M}^{d} \rightarrow \mathcal{N}^{d}$).}
  \label{fig:2}
\end{figure}

In general, a tangent map is a generalization of the function differentiation, and a tangent map is a linear mapping between two tangent spaces (shown in \figref{fig:2}). To be specific, given  $F: \mathcal{M}^{d} \rightarrow \mathcal{N}^{d}$, a mapping over surfaces, at each point $\mathbf{x} \in \mathcal{M}^{d}$, the tangent map $F^{*}$ is a linear transformation from the flat tangent space $\mathcal{T}_{\mathbf{x}} \mathcal{M}^{d}$ to the tangent space $\mathcal{T}_{F(\mathbf{x})} \mathcal{N}^{d}$.~\cite{o2006elementary}

\subsection{Graph Neural Network}
Graph neural networks can be viewed as an appropriate means to perform message passing between nodes~\cite{scarselli2008graph}. FMGNN uses a graph convolutional network proposed in~\cite{kipf2016semi}, where node embeddings are updated by aggregating adjacent node information. 

Based on general GCN message passing rule, and referring to the constant curvature graph convolution neural network, HGNN and HGCN, via exponential and logarithmic mapping, we can have the  general GCN message passing rules at layer $\ell$ for node $v_{i}$ on manifold $\hat{\mathcal{M}}$ ($\mathbb{E}, \mathbb{H}$ and $\mathbb{S}$) are shown as follows:

Feature transform function is :
\begin{equation}
  \label{r_ft}
  \mathbf{h}_{i}^{\ell, \hat{\mathcal{M}}}=W^{\ell} \exp _{\mathbf{x}^{\prime}}^{\hat{\mathcal{M}}}(\mathbf{v}_{i}^{\ell-1, \hat{\mathcal{M}}})+\mathbf{\lambda}^{\ell}.
\end{equation}

Neighborhood aggregation function is :
{\small
\begin{equation}
  \label{r_na}
  \mathbf{v}_{i}^{\ell, \hat{\mathcal{M}}}=\sigma( \log _{\mathbf{x}^{\prime}}^{\hat{\mathcal{M}}}(\mathbf{h}_{i}^{\ell, \hat{\mathcal{M}}}) +\sum_{v_{j} \in \mathcal{N}(v_{i})} w(v_{i}, v_{j}) \log _{\mathbf{x}^{\prime}}^{\hat{\mathcal{M}}}(\mathbf{h}_{j}^{\ell, \hat{\mathcal{M}}})),
\end{equation}
}
where $h$ represents the input and output of the middle layer of a network, $i$ for vertice $i$, $l$ for $l$-th layer, $M$ for different manifolds. Besides, $w(v_{i}, v_{j})$ is an aggregation weight that can be computed following the dot product and other operations of the respective Riemannian manifolds, $\mathbf{\lambda}^{\ell}$ is a bias parameters for layer $\ell$, and $\sigma$ is denoted as a non-linear activation function.

\section{Observations of Training GNN on Manifolds}
\begin{figure}[h]
  \centering
  \includegraphics[width=\linewidth]{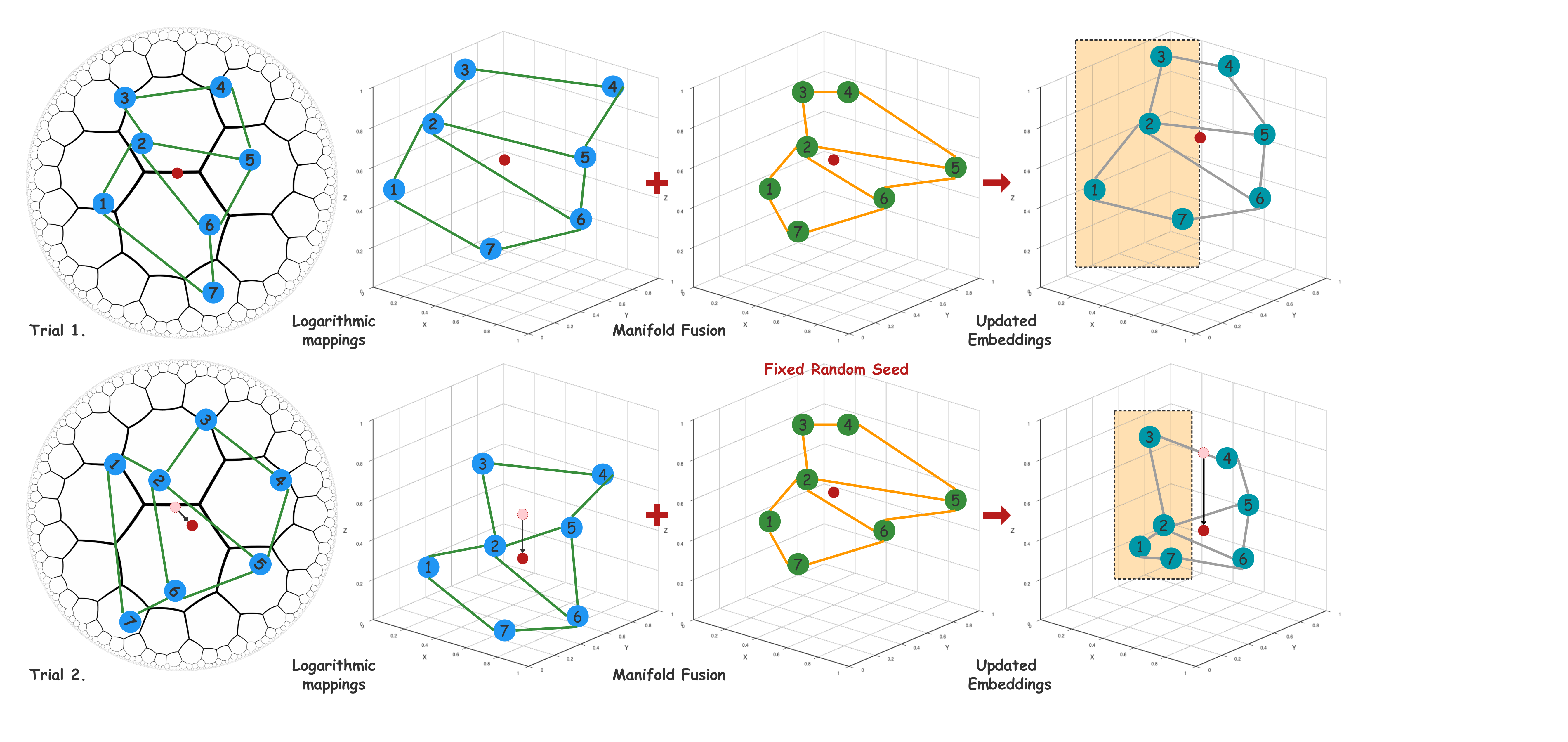}
  \caption{An example of combining embeddings from different spaces with random shifts. We fix the embedding on Euclidean space, obtain the hyperbolic embedding from two trails with different random seeds, and compare the results when combining the embeddings. 
 Take the docker in orange as an example, from which we can see that nodes 2 and 3 with connection on the original graph has a greater distance than nodes 2 and 7 without direct connection. Such a combination can result in errors in downstream tasks. }
  \label{fig:offset}
\end{figure}

In this section, we present our observations that the inherent randomness in GNN training can result in embedding centroid shifts and task performance fluctuations. 
When training GNN,  randomness is introduced from initialization and stochastic gradient descent. Due to such randomness, the final output embeddings of GNNs can be overall shifted or distorted, though the relative mutual relations are still preserved. 
Such a natural phenomenon can become a crucial issue when combining graph embeddings from different spaces. An example can be found in \autoref{fig:offset}. 

We investigate the centroid offsets and performance fluctuations over the final embeddings obtained by repeatedly running the same training process with different randomly generated seeds for GCNs on the three-manifolds, GIL, Hybrid-GCN, and our proposed FMGNN. 
We design Hybrid-GCN to demonstrate the performance over directly combining the embedding from three different manifolds without interaction, which extends the HGCN framework to directly superimpose the vertex embeddings from the three-manifolds, inspired by the constant curvature graph neural network~\cite{Bachmann2020ConstantCG}. 
FMGNN is our proposed model, listed for comparison, whose details we will describe in the following sections. 
We train GCN, HGCN (GCN on Hyperbolic manifold), SGCN (GCN on spherical manifold), GIL, Hybrid-GCN, and FMGNN on the datasets of Cora, PubMed and CiteSeer, 
for $10$ times and compare the performance. 

\begin{table}[h]
\begin{tabular}{@{}c|cccc@{}}
\toprule
\textbf{Benchmark} & \textbf{Model} & \textbf{Scale} & \textbf{Offset} & \textbf{Norm. Offset} \\ \midrule
\multirow{5}{*}{\textbf{Cora}} & GCN & 0.58 & 83.16 & 144.48 \\
 & HGCN & 0.48 & 31.20 & 65.49 \\
 & SGCN & 0.51 & 32.21 & 63.44 \\
 & Hybird-GCN & 1.30 & 393.05 & 301.60 \\ 
 & GIL & 1.32 & 215.22 & 162.27 \\ 
 & \textbf{FMGNN} & \textbf{1.60} & \textbf{11.14} & \textbf{6.97} \\
 \midrule
\multirow{5}{*}{\textbf{PubMed}} & GCN & 0.79 & 112.38 & 141.59 \\
 & HGCN & 0.69 & 51.37 & 74.71 \\
 & SGCN & 0.66 & 47.93 & 72.99 \\
 & Hybird-GCN & 1.49 & 531.12 & 356.45 \\ 
 & GIL & 0.34 & 161.18 & 475.32 \\ 
 & \textbf{FMGNN} & \textbf{1.89} & \textbf{14.93} & \textbf{7.91} \\
 \midrule
\multirow{5}{*}{\textbf{CiteSeer}} & GCN & 0.43 & 74.78 & 173.54 \\
 & HGCN & 0.25 & 22.80 & 90.56 \\
 & SGCN & 0.28 & 25.96 & 92.96 \\
 & Hybird-GCN & 2.58 & 414.93 & 160.82 \\
 & GIL & 0.25 & 207.75 & 848.54 \\ 
 & \textbf{FMGNN} & \textbf{3.20} & \textbf{11.75} & \textbf{3.67} \\
 \bottomrule
\end{tabular}%
\caption{Comparison of Models' Centroid Offsets.}
\label{tab:offset}
\end{table}

\subsection{Centroid Offsets}


For centroid offsets, we analyze the random coordinate shifts and distortions of embeddings. 
Given graph $G(V,E)$, 
on all the models, the graph embeddings $X = \{x_{i}| v_i \in V\}$ are obtained after each training, and we use
the average offsets among vertex embedding centroids $c = \frac{\sum_{v_i \in V} x_{i}}{|V|}$  of different training times for the same model to measure the shifts. 
Thus, the centroid offset is defined as  
\begin{equation}
    offset = \frac{2}{T(T-1)} \sum_{ i \neq j}d(c_{i}, c_{j}),
\end{equation}
where $c_i$ is the centroid on the $i$th run, $T=10$ is the number of training times, and $d(\cdot)$ is the distance function on respective spaces. 
Considering that the training results can have different scales, to make a fair comparison and reflect topology difference, 
we measure the scales of the graph embeddings by calculating the average distance between  vertices and their centroids: 
\begin{equation}
    scale = \frac{1}{|V|} \Sigma_{v_i \in V}d(x_{i},c).
\end{equation}
Thereafter, the centroid shift can be normalized by scale 
as 
\begin{equation}
    Norm.~Offset=\frac{offset}{scale}. 
\end{equation}
 The results can be found in \autoref{tab:offset}. 
We can see that GCN, HGCN, and SGCN have the same order of magnitude on normalized centroid offsets. If we combine the embedding results of different spaces as Hybrid-GCN or interact between manifolds without proper alignment, the normalized shifts can be accumulated from each space and become even greater. 
Such offset can result in variance and fluctuation in the performance, as we will show in the following subsection. 
Our proposed model has the most minor offsets.



\subsection{Performance Fluctuation}

We proceed to describe our observations on the performance fluctuation of models combining graph embeddings from different manifolds, namely Hybrid-GCN and GIL, to show that mixing such different manifold embeddings without proper alignment, can also result in performance variances, indicated by the large centroid offsets.  

\begin{figure}[h]
    \centering
    \includegraphics[width=0.8\linewidth]{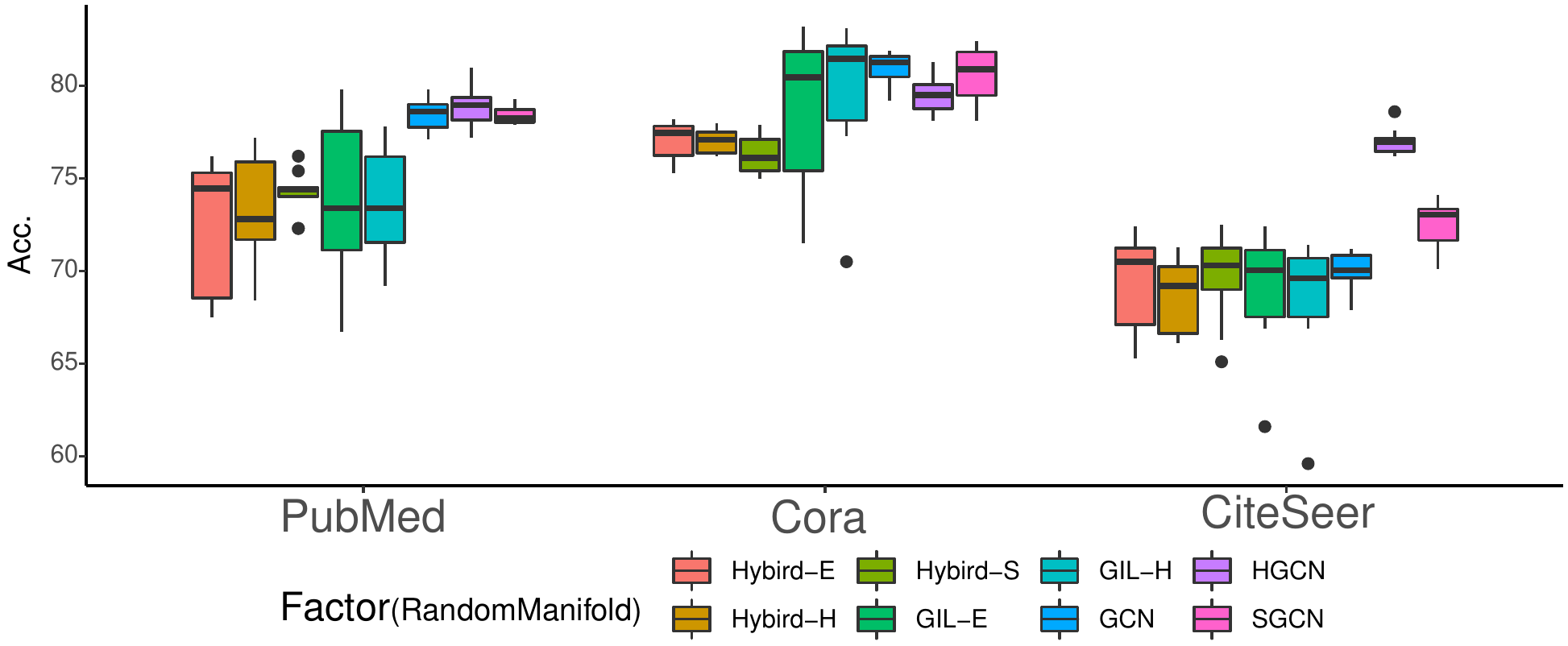}
    \caption{The performance fluctuations of the models on node classification task, where Hybird-$X$ denotes that we change the random seed of embedding training on $X$ manifold of the Hybird-GCN, and the same goes for GIL-$X$.}
    \label{fig:tentrain}
\end{figure}

For GIL and Hybrid-GCN, we choose to change embedding training only on one manifold, each at a time and remain the embeddings of the other manifolds unchanged.  
We observe the performance fluctuation introduced by the random shift of manifold embedding training with changed randomly selected seeds. 
The performance measured by the accuracy of node classification over the three datasets is demonstrated in \autoref{fig:tentrain}. 
For results in each dataset, the three rightmost boxplots are the performance variances of GCN, HGCN, and SGCN, which are used as benchmarks for the variances of GCN on the three different manifolds, while the results for Hybrid-GCN are shown on the leftmost and those for GIL are in the middle. 
We can see that for both Hybrid-GCN and GIL, the performance variances are more significant than the variances of GCN on any manifold. 
This indicates that simplifying combining embeddings from different manifolds without considering the proper alignment, 
the random shift and distortion of graph embedding on each manifold can be enlarged when combining the different embedding and result in greater performance fluctuation. 

Therefore, it is crucial to consider the random shifts of embedding on different manifolds on the embedding fusion and provide interaction and alignment between embeddings during training on all the manifolds 
if we choose to combine  embeddings on different manifolds to approximate the best-fit space for each vertex and enhance the performance. 
Now we present our proposed work FMGNN, which fusing different manifold embeddings with alignment during training and extract the mutual relation information on different manifolds by distances to the geometric coresets. 
\section{The Fused Manifold Graph Neural Network}
\label{sec:fmgnn}
Based on the observation above, FMGNN incorporates different manifolds information using a feature fusion mechanism, and converts coordinates on three manifolds into low-dimensional representations through geometric coreset based centroid mapping. We present our method in details as follows.

\subsection{Fused Manifold Aggregation}

FMGNN fuse the embedding from each manifold by interaction and alignment of the embeddings. 
First, according to \eqnref{r_ft}, we execute the feature transform function to obtain the node representation of each manifold, which is denoted as $\mathbf{h}_{i}^{\ell, \mathbb{E}}, \mathbf{h}_{i}^{\ell, \mathbb{H}}$ and $\mathbf{h}_{i}^{\ell, \mathbb{S}}$. 
According to the tangent mapping, if there is a smooth mapping between general manifolds, there is a linear mapping between the tangent spaces of the points before and after the mapping. 
Therefore, the representation vector of a point on one manifold can be mapped to another one via a selected linear mapping on the tangent spaces, and FMGNN takes advantage of this property. 
The linear mapping in FMGNN is trained from six trainable parameters denoted as $\lambda_{\mathbb{E} \rightarrow \mathbb{H}}, \lambda_{\mathbb{E} 
\rightarrow \mathbb{S}}, \lambda_{\mathbb{H}
\rightarrow \mathbb{E}}, \lambda_{\mathbb{H}  \rightarrow \mathbb{S}}, \lambda_{\mathbb{S}\rightarrow \mathbb{E}}$ and $\lambda_{\mathbb{S}\rightarrow \mathbb{H}}$, representing the weight of mapping between two manifolds.  

Taking hyperbolic manifold as an example, at first, the vectors on Euclidean manifold ($\mathbf{h}_{i}^{\ell, \mathbb{E}}$) are linearly mapped ($\lambda_{\mathbb{E} \rightarrow \mathbb{H}}$) to the tangent space of hyperbolic manifold, and further projected to the surface of hyperbolic manifold through exponential mapping, while the vectors on spherical space $\mathbf{h}_{i}^{\ell, \mathbb{S}}$ requires logarithmic functions to project the vectors on the surface to the tangent space of the sphere, through the linear mapping weight ($\lambda_{\mathbb{S} \rightarrow \mathbb{H}}$) to the tangent space of the hyperbolic manifold, and further projected into the surface of hyperbolic manifold via the exponential mapping. Finally, we use plain-vanilla $M\ddot obius$ addition \eqnref{m_add} to adjust the node embeddings and conduct neighborhood aggregation via Riemannian GCN aggregation function \eqnref{r_na}. 
Hence the feature fusion functions in a FMGNN layer is as follows:

\begin{equation}
    \mathbf{h}_{i}^{\ell, \mathbb{E}} = \mathbf{h}_{i}^{\ell - 1, \mathbb{E}} + \lambda_{\mathbb{H}\rightarrow \mathbb{E}} \log^{\mathbb{H}}\mathbf{h}_{i}^{\ell - 1, \mathbb{H}}
    + \lambda_{\mathbb{S}\rightarrow \mathbb{E}} \log^{\mathbb{S}}\mathbf{h}_{i}^{\ell - 1, \mathbb{S}},
\end{equation}

\begin{equation}
    \mathbf{h}_{i}^{\ell, \mathbb{H}} = \exp^{\mathbb{H}} \lambda_{\mathbb{E} \rightarrow \mathbb{H}} \mathbf{h}_{i}^{\ell - 1, \mathbb{E}} \oplus_{\mathbb{H}} \mathbf{h}_{i}^{\ell - 1, \mathbb{H}} 
    \oplus_{\mathbb{H}} \exp^{\mathbb{H}} \lambda_{\mathbb{S}\rightarrow \mathbb{H}} \log^{\mathbb{S}}\mathbf{h}_{i}^{\ell - 1, \mathbb{S}},
\end{equation}

\begin{equation}
    \mathbf{h}_{i}^{\ell, \mathbb{S}} = \exp^{\mathbb{S}} \lambda_{\mathbb{E} \rightarrow \mathbb{S}} \mathbf{h}_{i}^{\ell - 1, \mathbb{E}} \oplus_{\mathbb{S}} \mathbf{h}_{i}^{\ell - 1, \mathbb{S}}
    \oplus_{\mathbb{S}} \exp^{\mathbb{S}} \lambda_{\mathbb{H}\rightarrow \mathbb{S}} \log^{\mathbb{H}}\mathbf{h}_{i}^{\ell - 1, \mathbb{H}}.
\end{equation}

Such operations can be viewed as aligning embeddings of the same vertex into their weighted center on each manifold through interaction. 

Then on all the manifolds,  we conduct feature transformation and neighborhood aggregation via \eqref{r_ft} and \eqref{r_na}.
In this way, FMGNN executes GCN on three different Manifolds sharing the same weight matrix, fuses the information from other manifolds, and finally gets three groups of node representations, denoted as $A^{\mathbb{E}}=\{\alpha_{i}^{r}, 0 < i \leq |\mathcal{V}|\}$ for the Euclidean manifold, $A^{\mathbb{H}}=\{\beta_{i}^{r}, 0 < i \leq |\mathcal{V}|\}$ for the hyperbolic manifold, and $A^{\mathbb{S}}=\{\gamma_{i}^{r}, 0 < i \leq |\mathcal{V}|\}$ for the spherical manifolds, among which $\alpha$, $\beta$ and $\gamma$ denotes the final embeddings generated by three-manifolds.

\subsection{Geometric Coreset Based Centroid Mapping}
After obtaining embeddings from three manifolds, a challenge is that 
the parameter learning methods for regression and prediction are only applicable in the Euclidean manifold and the arithmetic rules in the three manifolds are different. 
In this work, we use distance to extract the mutual relation information between vertices on the three manifolds. 
We select a vertex set, called geometric coreset, as  ``landmarks'' on each manifold, 
 for each vertex calculate distances to all the vertices in the geometric coreset, 
 and use such distances as the new representation vector of the vertex on the manifold. 
 Such new coordinates are the mutual relations extracted from each manifold and can be applied to commonly used algebraic operations, such as addition and multiplication. 
We first define geometric coreset in FMGNN as follows:

\begin{definition}
  \textbf{Geometric Coresets in Manifolds}: Given a node set in a manifold $\mathcal{Z}=\{\zeta_{1},...,\zeta_{n}, \zeta_{i} \in \mathcal{M}\}$ and a sensitive quality function $Q(\mathcal{Z})$, a $m$-nodes geometric coreset $\mathcal{C} = \{c_{1}, c_{2}, ..., c_{m}\}, c_{i} \in \mathcal{M}$ is a subset of $\mathcal{Z}$ whose $Cost(\mathcal{C})$ can be used to approximate $Q(\mathcal{Z})$ for any set $\mathcal{Z}$ of $x$ points.
\end{definition}

The process of finding the geometric coreset is as follows. 
\begin{itemize}
    \item[1.]First, we find out a coreset $\mathcal{C} = \{c_{1}, c_{2}, ..., c_{|\mathcal{C}|}\}, c_{i} \in \mathcal{M}$, noting that the coresets are generated based on parameter-free methods and independent of node embeddings ($\mathcal{E}={e_{i}}, i \in |\mathcal{V}|$).
    \item[2.]Then, after project the coreset to the  manifolds by exponential mapping, the pairwise distance between $c_{i}$ and $e_{j}$ can be calculated based on the distance measurement of each manifold \eqref{m_add}, $\alpha_{i, j} = d(c_{i}, e_{j})$.
    \item[3.]Finally, we sum up all the distances $(\alpha_{0, j}, ..., \alpha_{|\mathcal{C}|, j}) \in \mathbb{R}^{|\mathcal{C}|}$ to represent position of $e_{j}$ mapping to the coreset, and $e_{j}$ is the node representation which can be learnt and updated in the GCN processes deployed on different Riemannian manifolds.
\end{itemize}

The dimensions of vertex representations are only decided by the number of vertices in the geometric coreset, which is constant in theory~\cite{har2004coresets} and small in empirical experiments. 
Therefore, the embedding of manifolds are further reduced to lower dimensions by the coreset.

In geometric coreset based centroid mapping, on different manifolds, we take advantage of the KMeans coreset \cite{har2004coresets}.
First, we generate a large number of nodes, $\mathcal{Z}=\{\zeta_{1},...,\zeta_{n}, \zeta_{i} \in \mathcal{E}\}$, represented by vectors with the same dimension with the embedding size in fused manifold aggregation process. 
Next, referring to~\cite{bachem2017scalable}, we perform KMeans clustering: randomly initialize $k$ cluster centroid points, and calculate the distance from each point in $\mathcal{Z}$ to these centroid points, then assign each data point to the cluster with the closest cluster centroid point and recalculate the centroid point for each cluster; we repeat the above steps until the algorithm converges. Finally, we have $k$ centroid point:
\begin{equation}
    \mathcal{P}=\{\rho_{1},...,\rho_{k}, \rho_{i} \in \mathcal{M}\}.
\end{equation}

After we obtain the geometric coreset and its corresponding vertex embeddings on the manifolds, we calculate distance between each vertex in $\mathcal{P}$ and each vertex in $\mathcal{G}$ with the embeddings from the output of fused manifold aggregation: 
\begin{equation}
    E^{\mathcal{M}} = d_{\mathcal{M}}(A^{\mathcal{M}}, \mathcal{C}_{\mathcal{M}}),
\end{equation}
where $\mathcal{M}$ denotes manifold $\mathbb{R}$, $\mathbb{H}$ and $\mathbb{S}$, and $E^{\mathcal{M}}$ is the centroid mapping's output from each manifold. Each of $E^{\mathcal{M}}$ is a $|\mathcal{C}|$-dimensional vector. The embedding obtained by geometric coreset based centroid mapping is a new feature of a node after training and passing related nodes' messages in the corresponding manifold. 
In this way, the embedding obtained by graph convolution of each manifold are transformed to new representations with the dimension decided by the geometric coresets, the mutual relations on different manifolds are extracted through the distances between vertices and such geometric coreset landmarks, and such extracted information can be applied to common algebraic operations. 

\subsection{Attention based Graph Embedding Fusion}
\begin{figure}[h]
  \centering
  \includegraphics[width=0.8\linewidth]{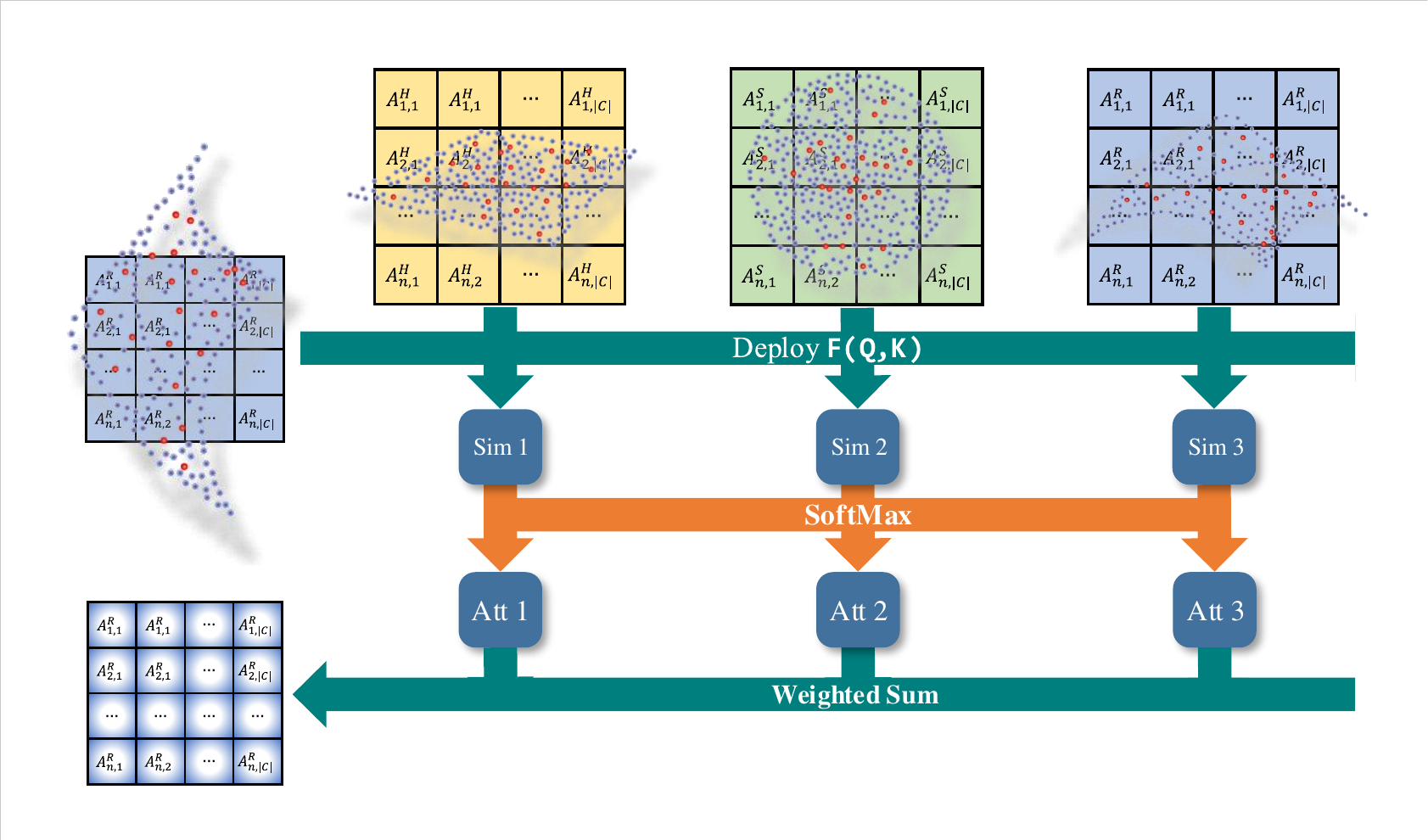}
  \caption{The self-attention mechanism in FMGNN. We take hyperbolic manifold as an example, we first \textit{query} the similarity with embeddings on other manifolds including hyperbolic manifold itself, and then adopt \textit{softmax} to gain the final weights.}
  \label{fig:att}
\end{figure}

Furthermore, we fuse these three embedding vectors via attention mechanism. 
In this module, FMGNN first to use the representation vector through geometric coreset of each point as the \textit{query} vector and such embeddings on all three manifolds (\textit{keys}) to calculate the similarity to obtain the weights, then use function $softmax$ to normalize these weights, and finally combine the weights with the corresponding embeddings on the manifold to perform weighted addition so as to generate the find graph embedding. The process is shown in Fig \ref{fig:att} and the equation of the attention is shown as follows.

{\small
\begin{equation}
E^{mm} = Attention(Q, K, V)=softmax\left(\frac{Q K^{T}}{\sqrt{d_{k}}}\right) V,
\end{equation}
}
where $d_{k}$ is the dimension of the embedding, equal to $|\mathcal{C}|$, the size of the coreset and $K$ is equal to $Q$. 

Finally, with the output, FMGNN is trained on node classification and link prediction tasks. For node classification, we map $E^{mm}=\{e_{i}^{mm}, 0 < i < |\mathcal{V}|\}$ to perform $softmax$ in order to find the most possible one from $\tau$ classes. The function is 
\begin{equation}
  p(e_{j}^{mm}) = softmax (\mathbf{w} (e_{1, j}^{mm},..., e_{|\mathcal{C}|, j}^{mm})), 0 < j < |\mathcal{V}|,
\end{equation}
where $\mathbf{w} \in \mathbb{R}^{\tau \times|\mathcal{C}|}$. For link prediction, we refer to the Fermi-Dirac method, a generalization of sigmoid, to compute probability for potential edges:
\begin{equation}
  p((i, j) \in \mathcal{V} \mid E_{i}^{mm}, E_{j}^{mm})=[e^{(d_{\mathbb{R}}(E_{i}^{mm}, E_{j}^{mm})^{2}-r) / t}+1]^{-1},
\end{equation}
where $r$ and $t$ are hyperparameters. We train on both node classification and link prediction tasks by minimizing the cross-entropy loss using negative sampling.

To sum up, FMGNN fuses embedding on different manifolds with interaction and alignment, 
aggregates information in the same space after the manifold fusion, 
and converts the embeddings on each manifold through geometric coreset based centroid mapping to exact mutual information. 

\section{Experiments}
In this section, we present and analyze our experiment results in detail. First, we describe real-life datasets and the baselines for comparison. After that, we compare these baselines with \textbf{FMGNN} on node classification (NC) and link prediction (LP) tasks. 
Moreover, we discuss the centroid offset of the FMGNN. Finally, we perform ablation experiments to show the necessity of geometric coresets and the contribution of different manifolds. 

\subsection{Experiments Setup}

\textbf{Benchmarks.} We use various open-sourced real-life graph datasets described in \tabref{tab:exp-bench}. 


\begin{table}[h]
  \begin{tabular}{lccccccc}
  \toprule
                    & \textbf{\#N.} & \textbf{\#Edge} & \textbf{\#Classes} & \textbf{\#Features} & \textbf{\#Dia.} & \textbf{\#Triangle} \\ \midrule
  Disease           & 1,044 & 1,043 & 2 & 1,000 & 10 & 0\\
  Airport           & 3,188 & 18,631 & 4 & 4 & 12 & 98,669 \\
  PubMed            & 19,717 & 44,327 & 3 & 500 & 18 & 12,520 \\
  CORA              & 2,708 & 5,429 & 7 & 1,433 & 19 & 1,630 \\
  CiteSeer          & 3,327 & 4,732 & 6 & 3,703 & 28 & 1,167 \\ 
  AmazonPhoto       & 13,381 & 119,043 & 8 & 745 & 11 & 717,400 \\ 
  Coauthor Phy.     & 34,493 & 245,778 & 5 & 8,415 & 17 & 468,550 \\ 
  \bottomrule
  \end{tabular}
  \caption{Statistics of the selected benchmarks.}
  \label{tab:exp-bench}
\end{table}

\begin{itemize}[leftmargin=0.8em]
  \item[1.] \textbf{Citation networks}. CORA, PubMed, and CiteSeer~\cite{kipf2016semi} are standard node classification and link prediction benchmarks, and the nodes are divided by research (sub)fields.
  \item[2.] \textbf{Disease propagation tree}~\cite{maulik2006advanced}. Disease, taken from HGCN, its labels of nodes is infected or not infected. 
  \item[3.] \textbf{Co-author networks}. Coauthor Physics is a co-authorship graph based on the Microsoft Academic Graph from the \href{http://kddcup2016.azurewebsites.net/}{KDD Cup 2016 challenge}. Its node labels are each author's most active fields of study.
  \item[4.] \textbf{Flight network}. Airport, origin from~\cite{zhang2018link}, is a transductive dataset updating by HGCN, whose nodes represent airports and edges represent the airline routes. The labels of nodes are the countries to which the airport belongs.
  \item[5.] \textbf{Goods co-consume network}. AmazonPhoto~\cite{shchur2018pitfalls, mcauley2015image} is a segment of the Amazon co-purchase graph, where the product categories give node labels. 
\end{itemize}

\textbf{Baselines.} We compare the performance of FMGNN with baselines, including shallow and neural network-based methods, GNNs in the Euclidean space, and GNNs in the hyperbolic space. For neural network-based methods, we choose MLP and HNN~\cite{ganea2018hyperbolic}. For GNNs in the Euclidean space, GCN~\cite{kipf2016semi}, GAT~\cite{velivckovic2017graph}, GraphSAGE~\cite{hamilton2017inductive} and SGC~\cite{wu2019simplifying} are taken into comparison. When it comes to GNNs in hyperbolic space, HGCN~\cite{chami2019hyperbolic}, HGNN~\cite{liu2019hyperbolic}, and H2H-GCN~\cite{Dai_2021_CVPR} are the most commonly adopted baselines. Besides, we also take GIL~\cite{Zhu2020GraphGI} into comparison as a baseline equipped with the embedding interaction only on the final outputs of Euclidean and hyperbolic manifolds. 

\textbf{Settings.} Referring to the HGCN, in the LP task, we use the random split point 85\%\//5\%\//10\% as the training sets, validation sets, and test sets. Moreover, in the NC task, we divide the disease with 30\%\//10\%\//60\% and 70\%\//15\%\//15\% for the airport, as for CORA, PubMed, and CiteSeer, we follow the division ratio in~\cite{kipf2016semi}, and for amazon photo and coauthor, we follow the division on the GNN benchmark~\cite{shchur2018pitfalls}. Following the HGCN and H2H-GCN, we evaluate link prediction by measuring the area under the ROC curve on the test set and evaluate node classification by measuring the F1 score, apart from CORA, CiteSeer, PubMed, where the performance is measured by accuracy. 

\subsection{Results}

\begin{table}[h]
  \caption{ROC AUC for Link Prediction (LP), and F1 score (DISEASE, binary class) and accuracy (the others, multi-class) for Node Classification (NC) tasks.}
  \label{tab:exp-res}
  \resizebox{\linewidth}{!}{
  \begin{tabular}{@{}ccccccccccccccc@{}}
  \toprule
  \textbf{Benchmarks} & \multicolumn{2}{c}{\textbf{DISEASE}} & \multicolumn{2}{c}{\textbf{AIRPORT}} & \multicolumn{2}{c}{\textbf{PUBMED}} & \multicolumn{2}{c}{\textbf{CORA}} & \multicolumn{2}{c}{\textbf{CITESEER}} & \multicolumn{2}{c}{\textbf{AmazonPhoto}} & \multicolumn{2}{c}{\textbf{Coauthor-Physics}} \\ \midrule
  \textbf{Hyperbolicity $\delta$} & \multicolumn{2}{c}{\textbf{0}} & \multicolumn{2}{c}{\textbf{1}} & \multicolumn{2}{c}{\textbf{2.5}} & \multicolumn{2}{c}{\textbf{3.0}} & \multicolumn{2}{c}{\textbf{4.5}} & \multicolumn{2}{c}{\textbf{2.5}} & \multicolumn{2}{c}{\textbf{2.5}} \\ \midrule
  \textbf{Task} & \textbf{LP} & \textbf{NC} & \textbf{LP} & \textbf{NC} & \textbf{LP} & \textbf{NC} & \textbf{LP} & \textbf{NC} & \textbf{LP} & \textbf{NC} & \textbf{LP} & \textbf{NC} & \textbf{LP} & \textbf{NC} \\ \midrule

  \textbf{MLP} & 63.6 $\pm$ 0.6 & 28.8 $\pm$ 2.5 & 89.8 $\pm$ 0.5 & 68.6 $\pm$ 0.6 & 84.1 $\pm$ 0.9 & 72.4 $\pm$ 0.2 & 83.1 $\pm$ 0.5 & 51.5 $\pm$ 1.0 & 86.3 $\pm$ 0.0 & 59.7 $\pm$ 0.0 & 69.1 $\pm$ 0.7 & 72.3  $\pm$ 1.8 & 68.8 $\pm$ 1.2 & 78.1 $\pm$ 3.1 \\
  \textbf{HNN} & 75.1 $\pm$ 0.3 & 41.0 $\pm$ 1.8 & 90.8 $\pm$ 0.2 & 80.5 $\pm$ 0.5 & 94.9 $\pm$ 0.1 & 69.8 $\pm$ 0.4 & 89.0 $\pm$ 0.1 & 54.6 $\pm$ 0.4 & 87.8 $\pm$ 0.0 & 77.2 $\pm$ 0.0 & 79.3 $\pm$ 4.2 & 88.7 $\pm$ 0.1 & 81.2 $\pm$ 0.8 & 87.7 $\pm$ 2.2 \\ \midrule
  
  \textbf{GCN} & 64.7 $\pm$ 0.5 & 69.7 $\pm$ 0.4 & 89.3 $\pm$ 0.4 & 81.4 $\pm$ 0.6 & 91.1 $\pm$ 0.5 & 79.0 $\pm$ 0.2 & 90.4 $\pm$ 0.2 & 81.5 $\pm$ 0.3 & 91.1 $\pm$ 0.0 & 70.3 $\pm$ 0.0 & 88.6 $\pm$ 0.6 & 91.2 $\pm$ 1.2 & 89.3 $\pm$ 3.1 & 92.8 $\pm$ 0.5 \\
  \textbf{GAT} & 69.8 $\pm$ 0.3 & 70.4 $\pm$ 0.4 & 90.5 $\pm$ 0.3 & 81.5 $\pm$ 0.3 & 91.2 $\pm$ 0.1 & 79.0 $\pm$ 0.3 & 93.7 $\pm$ 0.1 & 83.0 $\pm$ 0.7 & 91.2 $\pm$ 0.1 & 72.5 $\pm$ 0.7 & 90.0 $\pm$ 5.2 & 85.7 $\pm$ 20.3 & 92.0 $\pm$ 4.1 & 92.5 $\pm$ 0.9 \\
  \textbf{SAGE} & 65.9 $\pm$ 0.3 & 69.1 $\pm$ 0.6 & 90.4 $\pm$ 0.5 & 82.1 $\pm$ 0.5 & 86.2 $\pm$ 1.0 & 77.4 $\pm$ 2.2 & 85.5 $\pm$ 0.6 & 77.9 $\pm$ 2.4 & 87.9 $\pm$ 0.3 & 65.1 $\pm$ 0.1 & 89.9 $\pm$ 0.7 & 90.4 $\pm$ 1.3 & 92.1 $\pm$ 0.1 & 90.3 $\pm$ 0.8 \\
  \textbf{SGC} & 65.1 $\pm$ 0.2 & 69.5 $\pm$ 0.2 & 89.8 $\pm$ 0.3 & 80.6 $\pm$ 0.1 & 94.1 $\pm$ 0.0 & 78.9 $\pm$ 0.0 & 91.5 $\pm$ 0.1 & 81.0 $\pm$ 0.1 & 91.3 $\pm$ 0.1 & 71.4 $\pm$ 0.5 & 88.3 $\pm$ 0.2 & 83.7 $\pm$ 0.1 & 91.4 $\pm$ 0.3 & 92.0 $\pm$ 0.2 \\ \midrule
  \textbf{HGCN} & 90.8 $\pm$ 0.3 & 74.5 $\pm$ 0.9 & \underline{96.4 $\pm$ 0.1} & \textbf{91.6 $\pm$ 0.2} & 96.3 $\pm$ 0.0 & \underline{80.3 $\pm$ 0.3} & 92.9 $\pm$ 0.1 & 79.9 $\pm$ 0.2 & 93.9 $\pm$ 0.0 & \underline{77.6 $\pm$ 0.0} & \textbf{95.6 $\pm$ 0.2} & 90.0 $\pm$ 0.5 & 95.7 $\pm$ 0.6 & \underline{95.2 $\pm$ 0.3} \\
  \textbf{HGNN} & 90.6 $\pm$ 0.2 & 85.7 $\pm$ 0.8 & 95.8 $\pm$ 0.1 & 85.1 $\pm$ 0.3 & 94.1 $\pm$ 0.2 & 75.9 $\pm$ 0.1 & 96.1 $\pm$ 0.0 & 78.3 $\pm$ 0.2 & 97.4 $\pm$ 0.2 & 71.3 $\pm$ 0.1 & -- & -- & -- & -- \\
  \textbf{H2H-GCN} & \underline{97.0 $\pm$ 0.3} & 88.6 $\pm$ 1.7 & 96.4 $\pm$ 0.3 & 89.3 $\pm$ 0.5 & \underline{96.9 $\pm$ 0.0} & 79.9 $\pm$ 0.5 & 95.0 $\pm$ 0.0 & 82.8 $\pm$ 0.4 & 93.4 $\pm$ 0.2 & 65.7 $\pm$ 0.1 & \underline{93.4 $\pm$ 0.2} & \underline{91.3 $\pm$ 0.2} & \underline{96.1 $\pm$ 0.3} & 92.1 $\pm$ 0.4\\ \midrule
  
  \textbf{GIL} & \textbf{99.9 $\pm$ 0.0} & \underline{90.7 $\pm$ 0.4} & \textbf{98.7 $\pm$ 0.3} & \underline{91.3 $\pm$ 0.3} & 95.4 $\pm$ 0.1 & 78.9 $\pm$ 0.2 & \underline{98.2 $\pm$ 0.8} & \underline{83.6 $\pm$ 0.6} & \textbf{99.8 $\pm$ 0.1} & 72.9 $\pm$ 0.2 & 82.6 $\pm$ 2.3 & 84.3 $\pm$ 1.8 & 81.1 $\pm$ 5.1 & 83.0 $\pm$ 1.2 \\
  \textbf{FMGNN (Ours)} & 96.5 $\pm$ 0.4 & \textbf{90.9 $\pm$ 0.1} & 96.1 $\pm$ 0.4 & 88.1 $\pm$ 0.9 & \textbf{97.8 $\pm$ 0.1} & \textbf{80.4 $\pm$ 0.6} & \textbf{98.4 $\pm$ 0.2} & \textbf{83.9 $\pm$ 0.2} & \underline{98.1 $\pm$ 0.1} & \textbf{78.1 $\pm$ 0.8} & 89.8 $\pm$ 0.5 & \textbf{92.2 $\pm$ 0.3} & \textbf{96.3 $\pm$ 0.1} & \textbf{95.8 $\pm$ 0.2} \\ \bottomrule
  \end{tabular}}
\end{table}

\begin{figure}[t]
  \centering
  \subfigure[Comparison on NC Task]{
      \label{fig:radara}
      \includegraphics[width=0.4\linewidth]{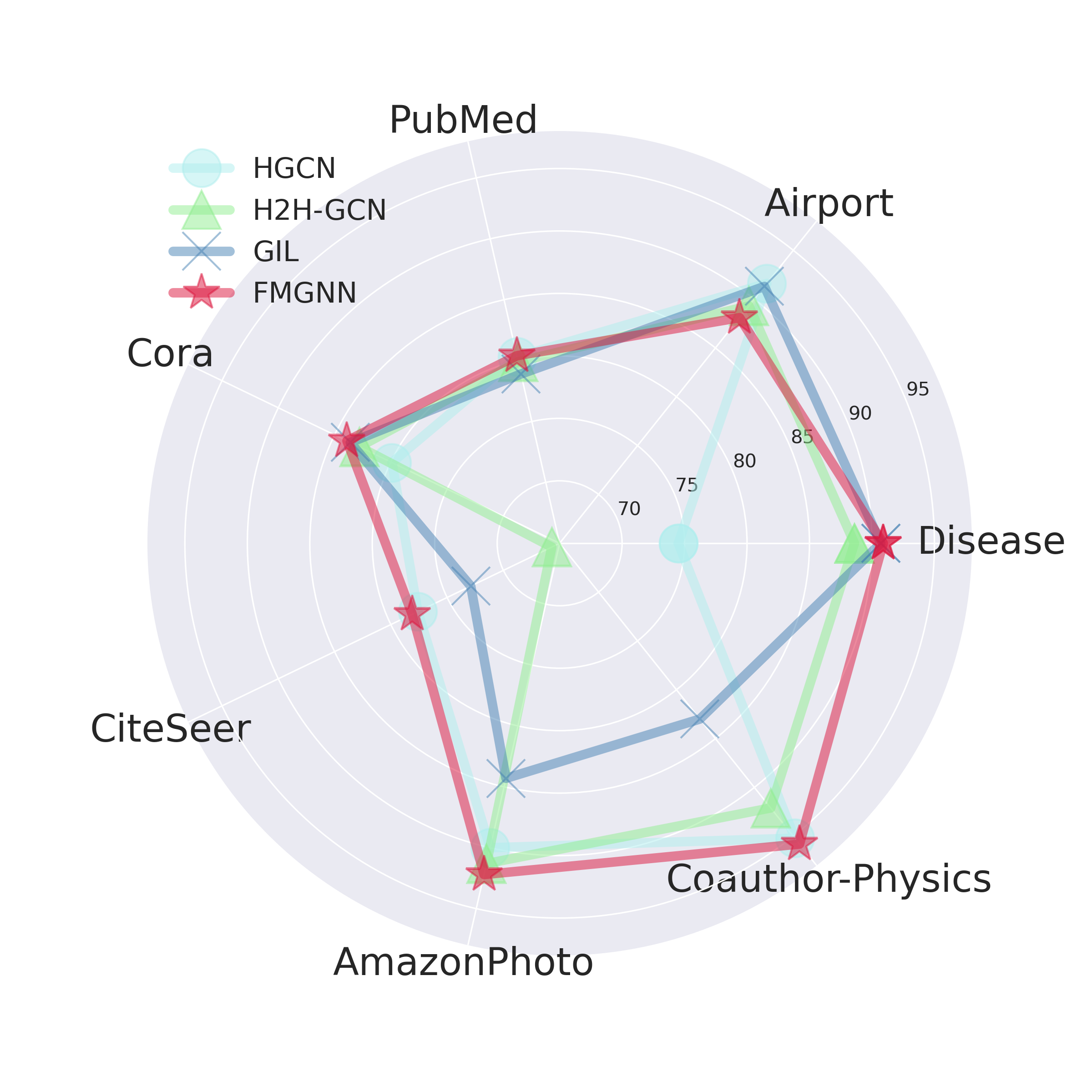}
  }
  \subfigure[Comparison on LP Task]{
      \label{fig:radarb}
      \includegraphics[width=0.4\linewidth]{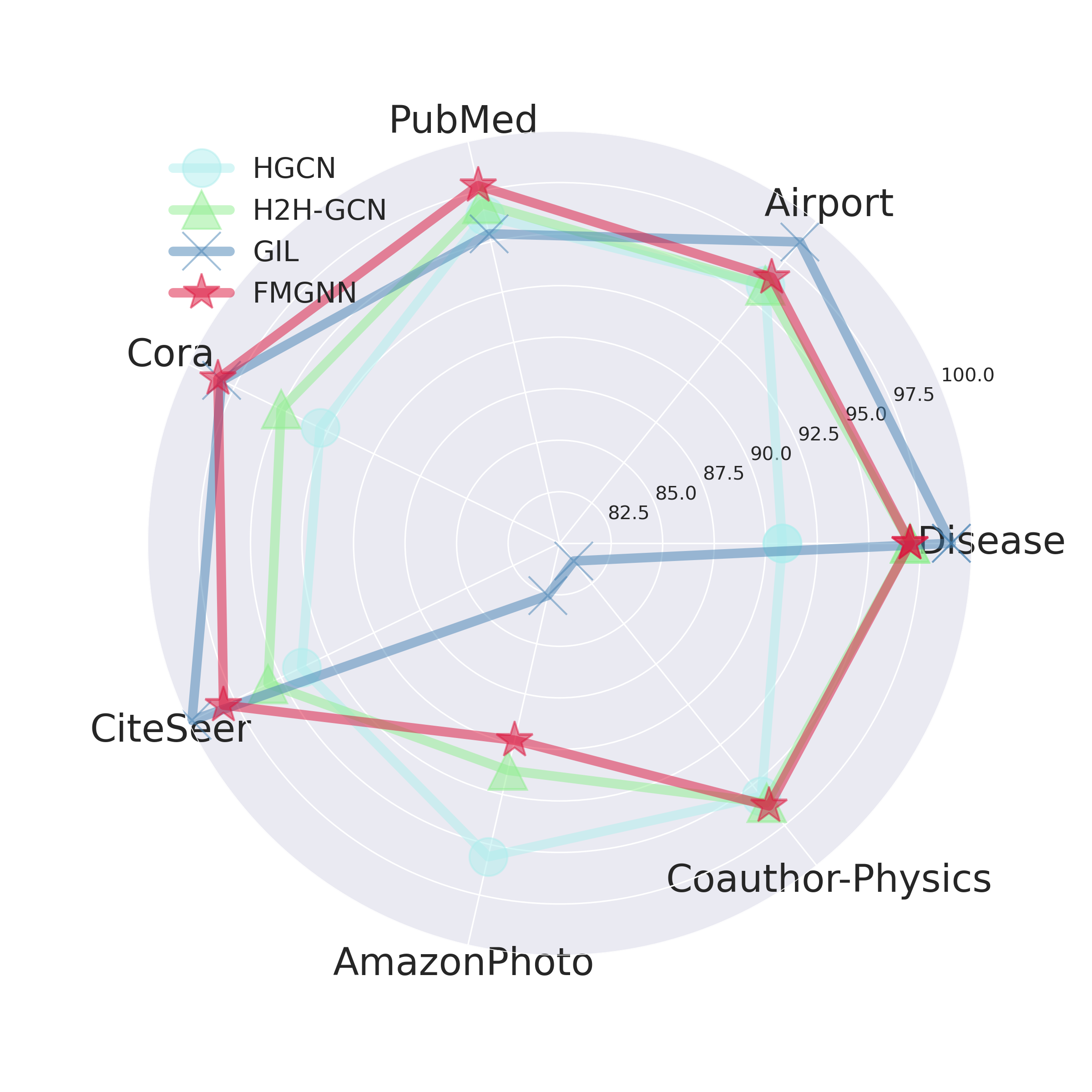}
  }
  \caption{The comparison of HGCN (color in Mint Tulip), H2H-GCN (color in Madang), GIL (color in Pigeon Post) and FMGNN (color in Mandy) on various benchmarks. In general, FMGNN performs well all-round.}
  \label{fig:radar}
\end{figure}

\tabref{tab:exp-res} 
summarizes the performance of FMGNN and the baseline methods on seven benchmarks, where the best results are shown in bold and the second best are shown with an underline. We ran five times and reported the average results and standard deviations. From the \tabref{tab:exp-bench} and \tabref{tab:exp-res} we can see that:
\begin{itemize}[leftmargin=0.8em]
    \item[1.] FMGNN outperforms the baselines or achieves similar results in classifying different types of nodes and link prediction between nodes.
    \item[2.] Compared to GIL, FMGNN has better performance on node classification tasks on six benchmarks and has more significant advantages on large-scale datasets such as AmazonPhoto and Coauthor Network on Physics due to the feature fusion and alignment before aggregation on each GCN layer.
    \item[3.] Compared to H2H-GCN, although our model achieves lower accuracy on the high hyperbolicity benchmark, the overall performance is close. On datasets such as Cora, Pubmed, and CiteSeer, the performance is significantly better than H2H-GCN. In particular, on a graph like CiteSeer has a large diameter and fewer triangle motifs, FMGNN performs better on graphs with sparsity.
    \item[4.] Our model achieves the best overall performance on all the datasets with different characteristics, compared to other baselines through the radar map in \autoref{fig:radar} (the line in red belongs to FMGNN).
\end{itemize}
We introduce the hyperbolicity measurements $\delta$ shown in \autoref{tab:exp-res}, which is originally proposed by Gromov~\cite{adcock2013tree}, to measure how much hierarchical structure a graph is inherent with, 
where the lower $\delta$ is, the more the graph is like a tree. 
Empirical results show that the more the graph is like a tree, the better performance the hyperbolic graph embedding can achieve: hyperbolic methods outperform the Euclidean methods in Disease, Airport. 
Our FMGNN outperforms HGCN and HGNN, meaning that FMGNN can benefit from embedding from the spherical manifolds and the Euclidean manifolds, even in the tree-like graphs. 

In the less tree-like graphs with higher $\delta$, such as PubMed, Cora, and CiteSeer, Euclidean manifold methods outperform the hyperbolic, and our FMGNN still achieves the best performance. 
Suppose a network diameter is large, such as Citeseer. In that case, models such as H2H-GCN and HGNN that are highly dependent on the ductility of hyperbolic space or even GIL which performs interact learning with Euclidean manifold have relatively low performance due to the fact that in the process of message passing, the distant nodes may be absorbed by accidental connections (weak correlations). 
Moreover, we also notice that GIL does not perform well on large-scale graph data with many triangles, AmazonPhoto and Coauthor on Physics, which indicates that GIL does not fit well for graphs with densely connected components. Besides, when processing graphs with triangles motifs, before the step of neighborhood aggregation fusing information from other manifolds is of significance.
Last but not least, the feature dimensions of FMGNN are always below $100$, and compared with HGNN, the number of parameters of FMGNN is significantly reduced, reducing the model's complexity. 

\subsection{Observation on Stability}

\begin{figure}[h]
    \centering
    \includegraphics[width=0.6\linewidth]{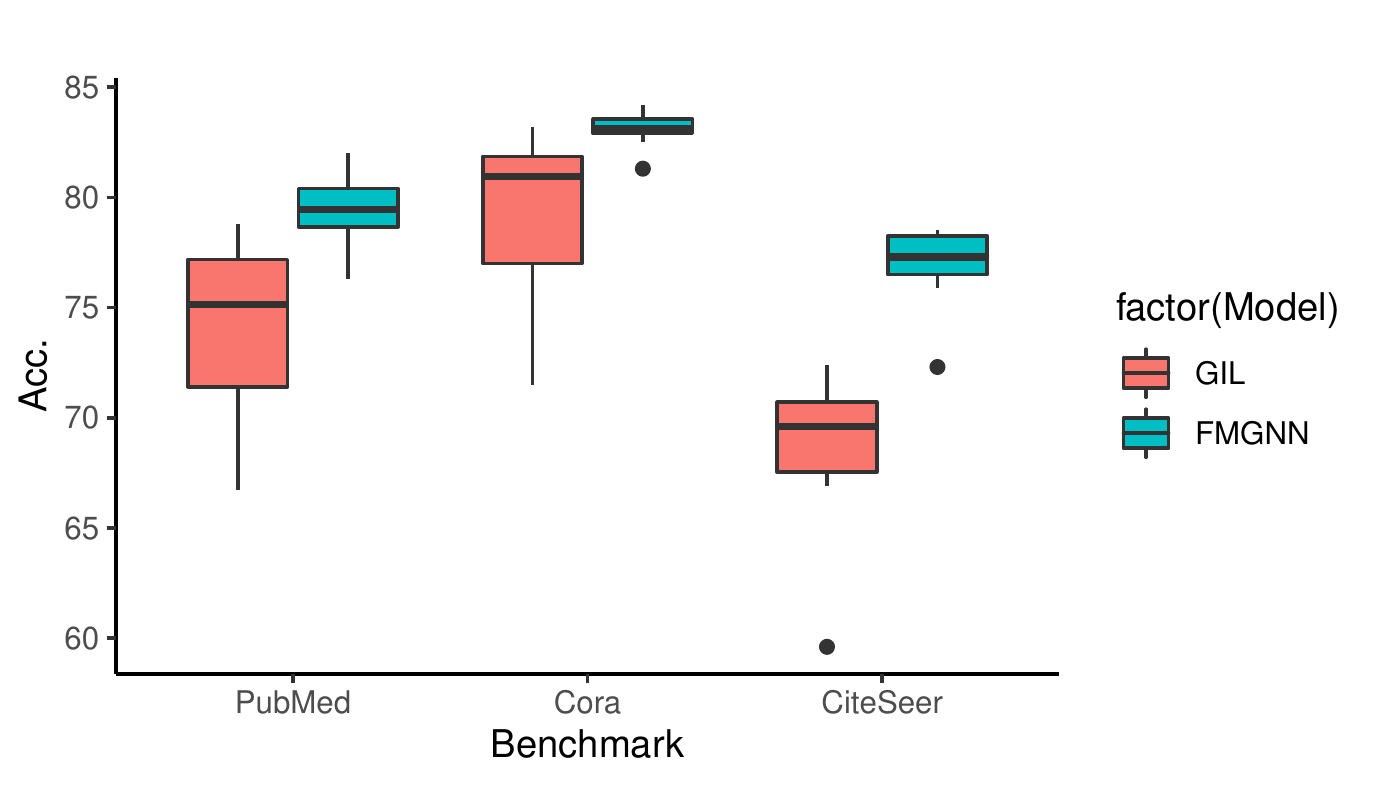}
    \caption{Fluctuations when training GIL and FMGNN.}
    \label{fig:cmpgil}
\end{figure}

Our model is more stable than the hybrid-GCN and GIL tested by the randomness. 
Compared with GIL, FMGNN interacts and aligns embeddings on different manifolds during training, and the mutual relation information on each manifold is preserved in the distances to the geometric coresets, while
GIL only executes the interaction learning after the aggregation. 
The performance shows that such alignment of FMGNN can reduce the random shifts as shown in Section 4 and performance fluctuation, which demonstrates the efficiency of FMGNN. 


\subsection{Ablation}

In order to better understand the role of each component of FMGNN, we conduct a detailed ablation experiment for FMGNN on CORA, PubMed, CiteSeer, AmazonPhoto, and Coauthor-Physics to observe the effects of geometric coreset-based centroid mapping and the necessity of using the three-manifolds. The  result is as follows.

\begin{figure}[h]
    \centering
    \includegraphics[width=0.6\linewidth]{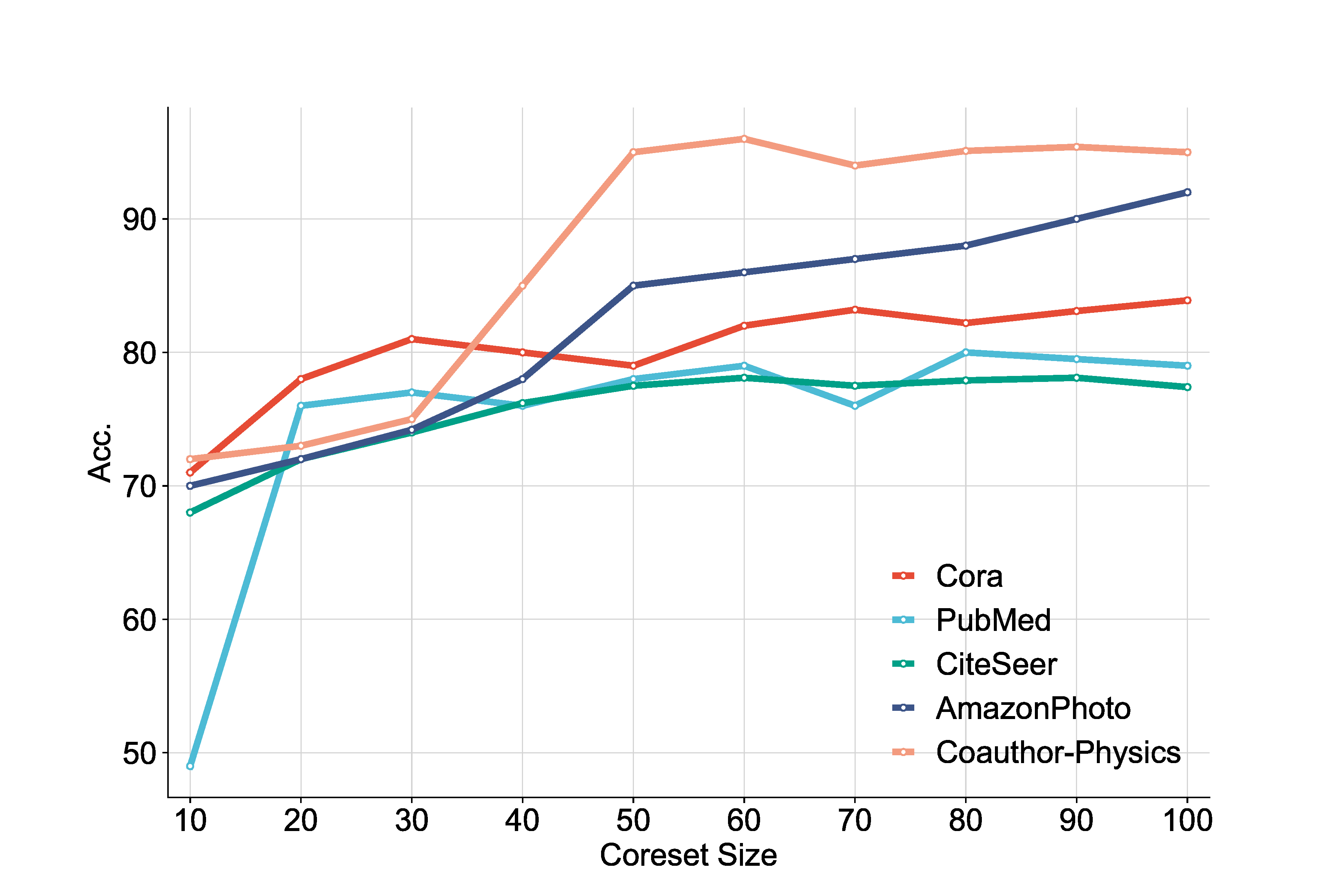}
    \caption{ACC of NC task distribution along with coreset size, where the X-axis denotes the step of the size of the coreset size, and the Y-axis denotes the ACC of FMGNN.}
    \label{fig:4}
\end{figure}

\textbf{The geometric coreset size.}  The FMGNN embedding dimension is decided by the size of the coresets in the three-manifold.  We observe the relation between the size of the coreset and the amount of the initial random nodes, and find that the dimension of the coreset is highly relevant to the size of the initially random nodes in each manifold. We set the size of the initially embedding of nodes on each manifold as 100, and we change the coreset size from 10 to 100, and the result is shown in \autoref{fig:4}.

We can find that our model performs best with 100 dimensions. At the same time, we notice that as the size of geometric coreset increases from 10 to 100, the accuracy does not rise simultaneously. The experiments show that there is a peak in the performance with respective to the size of coresets. 
What is more, it shows that the coreset reduces the dimension of the vertex embedding. Meanwhile, it also reflects that the larger the coreset, the more redundant information in the absorbed space. 

\textbf{Every manifold counts.} Given the introduction of GCN in the spherical manifold, we conduct parallel-space ablation experiments on our model. Experiments are performed on only one of the three spaces and discarding one. The experimental results over classic citation networks are shown in \tabref{every}.

\begin{table}[h]
\caption{FMGNN manifolds ablation experiments results.}
\label{every}
\centering
\begin{tabular}{c|ccc}
\hline
                                    & CORA & PubMed & CiteSeer      \\ \hline
$\mathbb{R}$                        & 81.3 & 79.1 & 72.8         \\
$\mathbb{H}$                        & 81.1 & 78.4 & 73.3          \\
$\mathbb{S}$                        & 81.8 & 78.9 & 72.0         \\
$\mathbb{R}+\mathbb{H}$             & 81.7 & 79.7 & 74.4         \\
$\mathbb{R}+\mathbb{S}$             & 82.5 & 79.9 & 73.8         \\
$\mathbb{S}+\mathbb{H}$             & 82.1 & 79.5 & 74.1         \\
$\mathbb{R}+\mathbb{H}+\mathbb{S}$  & \textbf{83.9} & \textbf{80.4} & \textbf{78.1} \\ \hline
\end{tabular}
\end{table}

As a result, centroid mapping in Euclidean manifold alone can achieve good results in PubMed, and hyperbolic manifold performs better in CiteSeer while spherical manifold suit CORA, which support the conclusion in~\cite{Bachmann2020ConstantCG}. It also shows that Euclidean, hyperbolic and spherical manifolds partially capture the primary information of the nodes on these three graphs. In addition, any combination of manifolds can increase the experimental result since the information each manifold captures is not the same. Finally, we can conclude that the information on the three spaces needs to be collected simultaneously to get better performance.

\section{Conclusion}
In this paper, we propose FMGNN, a graph neural network that fuses embeddings from  Euclidean, hyperbolic, and spherical manifolds with embedding interaction and alignment during training. 
Such embedding alignment can benefit from the representation capability of each manifold for its best-fit graph structures, while reducing the centroid offset and performance fluctuation during combining the embedding on different manifolds. 
We also propose the geometric coreset centroid mapping to fuse the mutual relation preserved in each manifold with low dimensions and high efficiency. 
As a result, we outperform the baseline models in most of the selected benchmarks, especially on node classification tasks.
In future, we would like to further exploit and propose the embedding mapping and interacting among different manifolds with conformal or length preserving insights with theoretical guarantees. 

\newpage
\bibliographystyle{ACM-Reference-Format}
\bibliography{ref}

\end{document}